\documentclass[runningheads]{llncs}
\usepackage[T1]{fontenc}
\usepackage{graphicx}
\usepackage{algorithm}
\usepackage{algpseudocode}
\usepackage{pifont}
\usepackage{hyperref}
\usepackage{booktabs}
\usepackage{amsfonts}
\usepackage{amssymb}
\begin{document}
\title{GNC-Pose: Geometry-Aware GNC-PnP for Monocular 6D Pose Estimation}
%
%
\author{Xiujin Liu\inst{1}\orcidID{0009-0000-7008-1270}}
\authorrunning{Xiujin Liu.}
%
\institute{University of Michigan
\email{jeanliu@umich.edu}\\
}
\maketitle              
\begin{abstract}
We present GNC-Pose, a fully learning-free monocular 6D pose estimation pipeline for textured objects. The proposed method combines rendering based initialization, geometry aware correspondence weighting, and robust Graduated Non-Convexity (GNC) optimization. Starting from coarse 2D-3D correspondences obtained from feature matching  alignment, our method builds on the Graduated Non-Convexity (GNC) principle and introduces a geometry aware weighting mechanism. Based on the 3D structural consistency of the model, the mechanism assigns robust confidence to per point as geometric priors. This geometric priors and weighting strategy bring stability to the estimation under outlier contamination. A final LM refinement further improve accuracy. We tested our method on The YCB Object and Model Set \cite{calli2015ycb}, despite the fact that it does not require learned features, training data, or category-specific priors, our work achieves competitive accuracy compared with both learning-based and learning-free methods, and offers a simple, robust, and practical solution for learning-free 6D pose estimation.

\end{abstract}
\section{Introduction}
Accurately detecting rigid objects and estimating their 6D poses from monocular RGB images alone is a crucial task in industrial manufacturing. Reliable pose estimation enables a wide range of downstream applications, including robotic grasping, manipulation, object tracking, and autonomous interaction in unstructured environments.\cite{xiang2017posecnn,wen2024foundationpose,di2021so,peng2019pvnet}

Although deep learning has greatly advanced the estimation of 6D poses in recent years, these methods still face several practical limitations. They typically require large amounts of annotated data, object-specific training, and careful domain adaptation to operate reliably in real-world scenarios. In contrast, learning-free pipelines have become a more attractive alternative in scenarios where training is infeasible, because of their generality, flexibility, and strong geometric interpretability. However, classic learning-free methods often rely on hand-crafted feature matching followed by classical PnP or RANSAC-PnP \cite{fischler1981random}, which struggle under real-world conditions. Feature correspondences are frequently contaminated by severe outliers, texture variable regions, occlusion, and self-symmetry, causing unstable optimization and inaccurate poses. More recent robust estimators based on the Graduated Non-Convexity (GNC) principle \cite{yang2020graduated} offer a mathematically grounded way to handle outliers. However, they are rarely integrated into a full end-to-end pipeline and typically ignore the underlying 3D structure clue of the object.

In this paper, we propose GNC-Pose, a monocular 6D object pose estimation pipeline for textured objects without the need for a GPU. After getting coarse correspondences from sparse feature matching between real RGB image and rendering-based synthetic image, we introduce a geometry aware weighting module that analyzes the 3D distribution of model points through voxelized clustering. This module assigns confidence scores to each point based on structural consistency, which effectively highlighting informative regions and suppressing ambiguous or symmetric surfaces. These confidence score guide a GNC-based PnP optimizer, in which the non-convexity parameter is gradually annealed to reject outliers and stabilize convergence. A final LM refinement further polishes the estimation.

To demostrate our method, we conduct comprehensive experiments on The YCB Object and Model Set \cite{calli2015ycb}. Based on the results, we demonstrate that GNC-Pose achieves competitive accuracy among both learning-based and learning-free methods while the pipeline requires no learned features, no training data, and no category-specific priors, making it deployable on any unseen textured object, as long as its CAD model is available. Extensive ablation studies demonstrate the effectiveness of each module.

Our contributions are summarized as follows:

1. A novel, unified, fully learning-free pipeline for monocular 6D pose estimation.

2. A geometry-aware weighting mechanism that assigns robust per-point confidence based on the 3D structural consistency of the object.

3. A GNC-based PnP optimization framework that effectively handles heavy outliers via gradual convexity annealing.
 
The code is available at \href{https://github.com/XiujinLiu/gnc-pose}{https://github.com/XiujinLiu/gnc-pose}

\section{Related Work}
\subsection{Learning-free 6D Pose Estimation}
Classical learning-free pipelines establish 6D poses by finding geometric consistency between 2D observations and 3D models. The Perspective-$n$-Point (PnP) problem serves as the mathematical foundation for this alignment, where solvers like EPnP \cite{lepetit2009ep} and DLS \cite{hesch2011direct} provide $O(n)$ solutions for estimating rotation and translation from sparse correspondences. To mitigate the impact of mismatched features, robust frameworks often integrate RANSAC \cite{fischler1981random} for iterative hypothesis generation or Iteratively Reweighted Least Squares (IRLS) \cite{holland1977robust} to modulate sample influence via continuous weights. For textured objects, hand-crafted descriptors such as SIFT \cite{lowe2004distinctive} and ORB \cite{rublee2011orb} are traditionally employed to build correspondence sets through template matching. More contemporary approaches, including SurfEmb \cite{haugaard2022surfemb} and RePose \cite{iwase2021repose}, explore the potential of synthetic depth maps and surface embeddings to refine the matching process. 

Despite their interpretability, these methods remain inherently vulnerable to "clustered outliers", which frequently lead standard PnP solvers to degenerate configurations or incorrect local minima.

\subsection{Learning-Based 6D Pose Estimation.}
Deep learning has enabled significant advances in 6D pose estimation from RGB or RGB-D inputs. Early direct regression approaches such as PoseCNN \cite{xiang2017posecnn} and CDPN \cite{li2019cdpn} focus on direct regression of pose parameters or disentangled coordinate prediction from inputs. To enhance localization precision, correspondence-based and refinement methods like PVNet \cite{peng2019pvnet}, GDR-Net \cite{wang2021gdr}, and CosyPose \cite{labbe2020cosypose} utilize dense feature learning, correspondence prediction, and multi-view fusion to achieve accuracy. More recent work further improves robustness and generalization, for example SO-Pose \cite{di2021so}, which exploits self-occlusion cues for direct 6D pose regression from RGB, and FoundationPose \cite{wen2024foundationpose}, which uses large-scale render-and-compare or foundation models to handle novel objects given only CAD models or a few reference images.

While achieving impressive accuracy, these models often require high-end GPU resources and substantial training data, making them difficult to deploy in embedded robotics environments or scenarios where only a raw CAD model is available for a new object.

\subsection{Robust Non-Convex Optimization and GNC.}
Robust estimation in the presence of extreme outlier ratios remains a fundamental challenge in spatial perception. Graduated Non-Convexity (GNC) \cite{yang2020graduated} has emerged as a powerful optimization paradigm that avoids the local minima issues typical of traditional M-estimators. By starting with a convex surrogate and gradually annealing towards a non-convex objective—such as the Geman--McClure \cite{geman1987statistical} loss—GNC enables stable convergence in highly noisy environments. This strategy has demonstrated state-of-the-art performance in tasks like certifiably robust geometric perception \cite{yang2020one} and fast point cloud registration as seen in TEASER \cite{yang2020teaser}. 

Our work builds upon this principle but introduces a novel structural prior; by integrating geometry-aware reweighting with the GNC-PnP solver, we provide the necessary spatial regularization that classical pipelines lack, ensuring robustness even when correspondences are sparse or heavily contaminated.

\subsection{Gaps}
Several gaps remain. While deep learning methods have achieved impressive accuracy, these approaches require extensive training data, object-specific fine-tuning, or large foundation models. They often generalize poorly to unseen objects, rely on high-quality meshes, or demand GPU resources that limit deployment in robotics settings. Moreover, their behavior is less transparent, making debugging and failure analysis more difficult. Classical 6D pose estimation pipelines rely on geometric cues such as feature matching, PnP, ICP, or multi-view rendering. These methods have the advantage of being learning-free, fully interpretable, and can be applied to any new object as long as a CAD model is available. However, their performance is often affected by imperfect correspondences, symmetries, occlusion, and noise in either the image or the CAD model. Without strong regularization or learned priors, classical pipelines tend to produce unstable inliers, degenerate PnP updates, or convergence to incorrect modes under ambiguity.

Bridging this gap requires methods that preserve the generality and interpretability while achieving robustness. Our work attempts to address this challenge by incorporating geometry-aware weighting and a graduated non-convexity PnP optimization mechanism, enabling a fully learning-free pipeline to remain stable even under texture-variable regions, mismatching, and incomplete or inaccurate CAD geometry.

\section{Methodology}
\subsection{Architecture}

\begin{figure}[t]
  \centering
  \includegraphics[width=\linewidth]{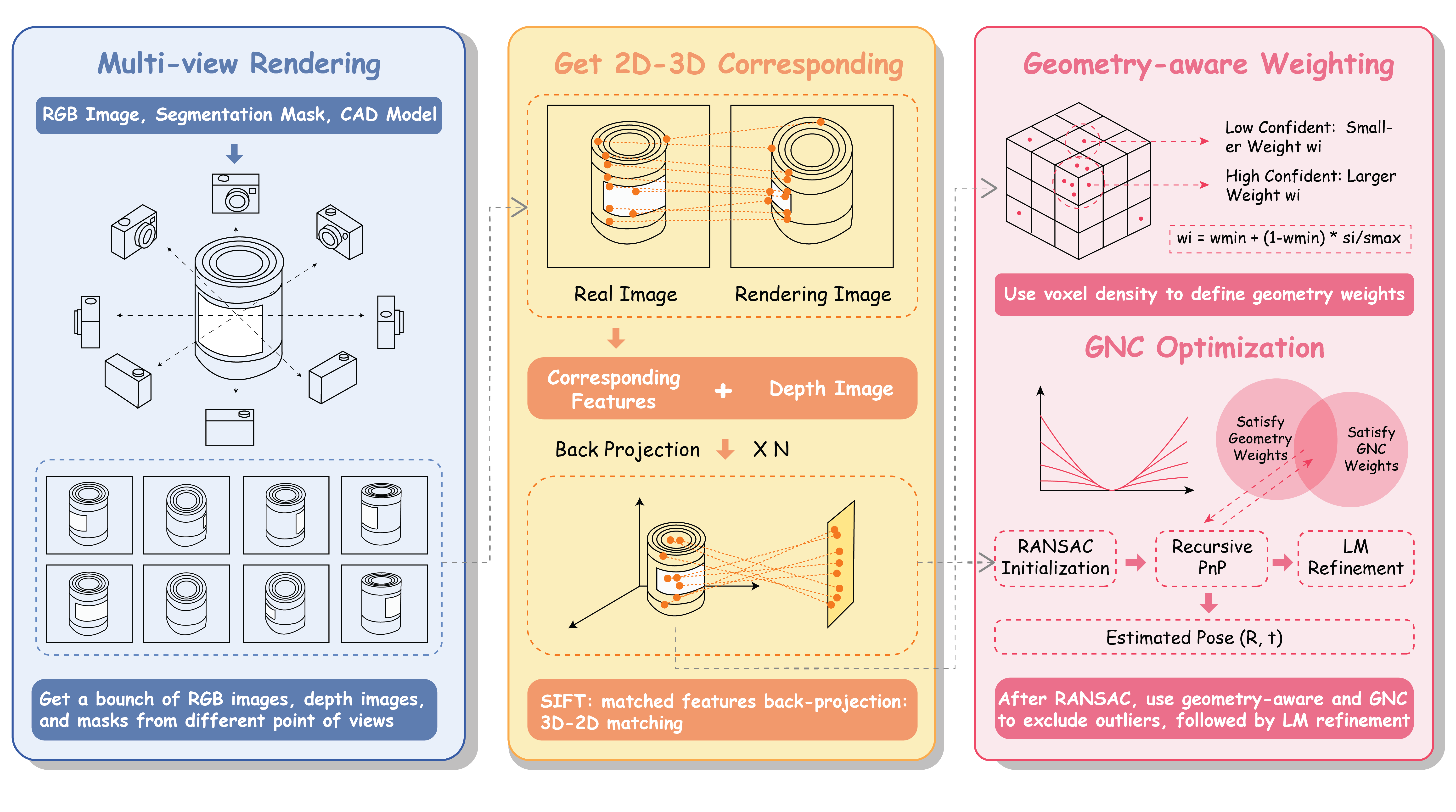}
  \caption{\textbf{Overview of the GNC-Pose pipeline.} We present a fully learning-free 6D pose estimation pipeline for textured objects based on robust GNC-PnP optimization with geometry aware reweighting. The workflow begins with Multi-view Rendering (left), which generates synthetic RGB-D templates from CAD models to establish a viewpoint reference. In the 2D-3D Correspondence stage (center), features are matched between real and rendered images and then back-projected into 3D space. Finally, the Robust Optimization stage (right) employs a Geometry-aware Weighting mechanism to assign per-point confidence based on voxel density, which guides the Graduated Non-Convexity (GNC) optimization to effectively suppress outliers and achieve precise 6D pose $(R, t)$ estimation.}
  \label{fig:pipeline}
\end{figure}

The overall architecture of GNC-Pose is as shown in Figure \ref{fig:pipeline}, Given an input RGB image and the CAD model of the target object, the pipeline begins by performing a multi-view rendering sweep around the object model. By rendering the CAD model across an efficiently sampled set of azimuth and elevation angles, we obtain a library of synthetic images; by comparing synthetic images with real image, we get matched key points and rich geometric clues. 
To mitigate residual noise, GNC-Pose applies a geometry-aware weighting module, which voxelizes the 3D model and evaluates the structural consistency of each region. This produces weights to each point that emphasize discriminative geometric regions and down-weight ambiguous or symmetric surfaces. The weighted correspondences are then fed into a GNC-based PnP optimizer, where the non-convexity parameter is gradually annealed to reject outliers and stabilize convergence. After GNC converges, a final LM-based refinement further \cite{levenberg1944method} sharpens the pose estimate.

\subsection{Geometry Aware Weighting Module}

This module is specifically designed to address the limitations of imperfect correspondences and geometric ambiguities faced by classical learning-free pipelines. As discussed above, traditional learning-free methods typically rely on RANSAC or standard PnP solvers, which often struggle when faced with "clustered outliers" caused by incorrect matches that are spatially concentrated due to self-symmetry or repetitive textures. While standard estimators treat all matched features with equal initial importance, they are highly susceptible to such "clustered outliers" or noises. To bridge this gap, GNC-Pose introduces a geometry-aware weighting module that leverages the 3D structural consistency of the object as a geometry prior. By reweighting correspondences based on their local density in the object coordinate space, the pipeline can effectively distinguish between stable, geometrically informative points and isolated, noisy mismatches, providing the necessary regularization that classical methods typically lack. And finally improve robustness under noisy correspondence generation.

To be more specific, given a set of $N$ sparse 2D--3D correspondences $\mathcal{P} = \{(x_i, X_i)\}_{i=1}^N$, where $x_i \in \mathbb{R}^2$ denotes the 2D feature coordinates in the image plane and $X_i \in \mathbb{R}^3$ represents the corresponding 3D coordinates in the object coordinate system, GNC-Pose constructs geometry-aware confidence weights by analyzing the spatial distribution and density of these 3D points.

Each 3D point $X_i$ is first mapped into a discrete voxel grid with a resolution of $v$ (the voxel size) through the quantization function:\begin{equation}g_i = \lfloor X_i / v \rfloor \in \mathbb{Z}^3\end{equation}where $g_i$ serves as a discrete spatial index. Points sharing the same index are grouped into a local geometric cluster $\mathcal{C}_{g_i} = \{ j \mid g_j = g_i \}$. The support count $s_i = |\mathcal{C}_{g_i}|$ acts as a discrete estimator of local matched-point density, which approximates the continuous density $\rho(X_i) v^3$ for a sufficiently small $v$.

We then normalize these support values by the global maximum $s_{\max} = \max_j s_j$ to obtain the relative density $\hat{s}_i = s_i / s_{\max} \in [0, 1]$. These are transformed into continuous geometry-aware weights:
\begin{equation}
w_i^{\mathrm{geom}} = w_{\min} + (1 - w_{\min}) \hat{s}i
\end{equation}
ensuring $w_i^{\mathrm{geom}} \in [w{\min}, 1]$. Here, $w_{\min}$ is a lower-bound hyperparameter that prevents the total rejection of isolated points. These weights reflect the structural reliability of each correspondence: dense, geometrically consistent regions receive $w_i \approx 1$, while isolated or noisy correspondences are effectively suppressed.

The resulting weights are integrated into a robust Graduated Non-Convexity (GNC) framework. Given the reprojection error $r_i(R,t) = \|\pi(K(R X_i + t)) - x_i\|^2$, where $K$ is the camera intrinsic matrix and $(R, t)$ represents the 6D pose, we compute the Geman-McClure soft inlier score:
\begin{equation}
w_i^{\mathrm{gnc}}(\mu) = \frac{\mu^2}{r_i^2 + \mu^2}
\end{equation}
In this formulation, $\mu$ is the GNC control parameter that is gradually annealed to adjust the non-convexity of the cost function. Combined with the geometry-aware prior $w_i^{\mathrm{geom}}$, we define the active inlier set $\mathcal{I}(\mu)$ at each iteration as:
\begin{equation}
    \mathcal{I}(\mu) = \left\{ i \mid w_i^{\mathrm{gnc}}(\mu) > \tau_{\mathrm{gnc}} \land w_i^{\mathrm{geom}} > \tau_{\mathrm{geom}} \right\}
\end{equation}
where $\tau_{\mathrm{gnc}}$ and $\tau_{\mathrm{geom}}$ are fixed confidence thresholds. Rather than contributing to a weighted objective function, these weights serve as a dual-filter mechanism to prune outliers and stabilize the PnP solver during the optimization process.

\subsection{GNC-Based PnP with Geometry-Aware Inlier Selection}

To solve the non-convex optimization challenges and the sensitivity to high outlier ratios that often plague classical PnP solvers, GNC-Pose employs Inlier Selection mechanism in Graduated Non-Convexity (GNC) module. While the geometry-aware weights provide a robust spatial prior, the presence of heavy correspondence noise still creates a highly non-convex error landscape with numerous local minima. Traditional local optimizers or standard RANSAC-PnP often collapse under these conditions, especially when the initial pose is coarse. By incorporating GNC, we can systematically anneal the cost function from a nearly convex surrogate to the target robust M-estimator. This allows the pipeline to progressively "filter" out outliers in a mathematically grounded manner, ensuring stable convergence to the global optimum even when the initial matching is significantly contaminated.

Given the geometry-weighted correspondences from Sec. 3.2, we estimate the object pose by minimizing a robust reprojection objective. For each correspondence $(x_i, X_i)$, we define the squared reprojection residual:
\begin{equation}
r_i^2(R, t) = | \pi(K(R X_i + t)) - x_i |^2
\end{equation}
where $\pi(\cdot)$ denotes the camera projection function, $K$ is the camera intrinsic matrix, and $(R, t)$ represents the 6D pose (rotation and translation). We then adopt the Geman–McClure M-estimator, whose GNC surrogate takes the form:
\begin{equation}
\rho_\mu(r) = \frac{r^2}{r^2 + \mu^2}
\end{equation}
Here, $\mu > 0$ is the continuation parameter that controls the non-convexity of the loss. When $\mu$ is large, the penalty $\rho_\mu(r) \approx r/\mu$ is nearly convex and behaves similarly to a scaled $\ell_2$ loss; as $\mu \to 0$, it becomes a sharply redescending and approaches a truncated cost that saturates for large residuals.

The associated influence function:
\begin{equation}
    \psi_\mu(r) = \frac{\partial \rho_\mu(r)}{\partial r}
= \frac{\mu}{(r + \mu)^2}
\end{equation}
reveals that that the contribution of a sample vanishes as $r$ grows, and that decreasing $\mu$ makes this transition sharper. In the limit $\mu\to 0$, $\psi_\mu(r) \to 0$ for all $r>0$, i.e., only exact inliers retain non-zero influence.

GNC implements this continuation principle by optimizing a sequence of surrogate problems with gradually decreasing $\mu$. In optimization practice, instead of using $\psi_\mu(r)$ explicitly, we employ the standard Geman–McClure soft inlier score, for a fixed $\mu$, we define a GNC soft inlier score:
\begin{equation}
    w_i^{\mathrm{gnc}}(\mu)
= 
\frac{\mu^2}{\, r_i^2 + \mu^2 \,}
\end{equation}
which is monotonically decreasing in $r_i$ and and behaves as a smooth soft-thresholding function: $w_i^{\mathrm{gnc}}(\mu) \to 1$ when $r_i \ll \mu$ and $w_i^{\mathrm{gnc}}(\mu) \to 0$ when $r_i \gg \mu$. In a conventional IRLS interpretation, such weights would modulate each correspondence’s contribution to a weighted normal equation. Rather than using them as continuous weights, GNC-Pose combine them with our geometry-aware structural prior $w_i^{\mathrm{geom}}$ to form a correspondence-selection rule
$\mathcal{I}(\mu)
=
\left\{
i \;\middle|\;
w_i^{\mathrm{gnc}}(\mu) > \tau_{\mathrm{gnc}}
\;\land\;
w_i^{\mathrm{geom}} > \tau_{\mathrm{geom}}
\right\}$. This discrete selection mechanism mirrors the behavior of a redescending M-estimator whose effective shape sharpens as $\mu$ decreases: for large $\mu$, almost all correspondences satisfy the inequality and the objective behaves close to least squares; as $\mu$ becomes small, only points with both small residuals and reliable geometry survive, mimicking a redescending loss with hard outlier rejection.

Algorithmically, we alternate between updating the inlier set $\mathcal{I}(\mu)$ and solving an weighted PnP on this set. Starting from an initial pose obtained by \texttt{solvePnPRansac}, we compute residuals $r_i$ and set:
\begin{equation}
    \mu_0 = \kappa \cdot \mathrm{median}_i(r_i) + \varepsilon
\end{equation}
where $\kappa$ is a scaling factor that determines how close the initial GNC stage remains to a convex regime. A larger $\kappa$ yields a more conservative (nearly convex) initialization with broader soft inlier acceptance, while a smaller $\kappa$ initiates the optimization with sharper non-convexity and more aggressive outlier suppression. In practice, $\kappa$ is chosen in the range $[3,10]$ depending on the expected noise level of the initial correspondences.

We then iterate the following steps:  

(i) evaluate $w_i^{\mathrm{gnc}}(\mu)$ for all correspondences;  

(ii) form the inlier set $\mathcal{I}(\mu)$ using the GNC and geometry thresholds;  

(iii) if $|\mathcal{I}(\mu)|$ is above a minimum inlier count, update $(R,t)$ by running \texttt{solvePnP} with \texttt{SOLVEPNP\_ITERATIVE} on the restricted set;  

(iv) recompute residuals and decrease $\mu$ according to 
\begin{equation}
    \mu \leftarrow \max(\gamma \mu,\; \mu_{\mathrm{final}}),
\qquad 0 < \gamma < 1
\end{equation}

As $\mu$ is annealed, the inlier set monotonically contracts and the surrogate objective increasingly approximates the target non-convex M-estimator. When $\mu$ reaches its final value or convergence is detected, we perform a last local refinement using \texttt{solvePnPRefineLM} on the final inlier set. The complete procedure is summarized in Algorithm~\ref{alg:gncpnp}. 







\begin{algorithm}[t]
\caption{GNC-PnP with Geometry-Aware Inlier Selection}
\label{alg:gncpnp}
\begin{algorithmic}[1]

\Require 2D--3D correspondences $\{(x_i, X_i)\}$, intrinsics $K$, geometry weights $w_i^{\mathrm{geom}}$.
\Ensure Final pose $(R,t)$, inlier mask $\mathcal{I}$.

\Statex \textbf{Variables:} 
\Statex $\pi(\cdot)$: Camera projection function; 
\Statex $\mu$: GNC control parameter (non-convexity scale); 
\Statex $w_i^{\mathrm{gnc}}$: Robust Geman-McClure soft inlier score; 
\Statex $\gamma \in (0, 1)$: Annealing factor;
\Statex $\tau_{\mathrm{gnc}}, \tau_{\mathrm{geom}}$: Fixed selection thresholds.
\Statex \hrulefill

\State $(R,t) \gets$ \texttt{PnPRANSAC} \Comment{Initialization}
\State Compute $r_i = \|\pi(K(RX_i + t)) - x_i\|^2$ \Comment{Squared residuals}
\State $\mu \gets k \cdot \mathrm{median}(r_i) + \varepsilon$ \Comment{Initial scale}

\Repeat
    \State $w_i^{\mathrm{gnc}} \gets \mu^2 / (r_i^2 + \mu^2)$ \Comment{Update soft weights}
    \State $\mathcal{I} \gets \{ i \mid w_i^{\mathrm{gnc}} > \tau_{\mathrm{gnc}} \ \land\ w_i^{\mathrm{geom}} > \tau_{\mathrm{geom}} \}$ \Comment{Dual-filter selection}
    \If{$|\mathcal{I}| < \texttt{min\_inliers}$}
        \State \textbf{break}
    \EndIf
    \State $(R,t) \gets \texttt{solvePnP}(X_{\mathcal{I}}, x_{\mathcal{I}}; K, R, t)$ \Comment{Pose update on inliers}
    \State Recompute $r_i$
    \State $\mu \gets \max(\gamma\mu,\,\mu_{\mathrm{final}})$ \Comment{Gradual annealing}
\Until{$\mu \le \mu_{\mathrm{final}}$}

\State $(R,t) \gets \texttt{solvePnPRefineLM}(X_{\mathcal{I}}, x_{\mathcal{I}}; K, R, t)$ \Comment{Final precision polish}

\end{algorithmic}
\end{algorithm}

\section{Experiment} 
\subsection{Dataset}
We evaluate GNC-Pose on The YCB Object and Model Set \cite{calli2015ycb}. The YCB Object and Model Set is one of the most widely used benchmarks for 6D pose estimation, providing a curated collection of household objects with accurate, high-resolution 3D meshes. The objects span a broad spectrum of geometric and photometric difficulty—including textureless surfaces, specular materials, and repetitive patterns, making the dataset well suited for evaluating correspondence robustness and pose stability. 

In our experiments, we focus on a subset of 12 objects featuring distinctive textures and geometries. This selection allows us to specifically assess the performance of feature-based correspondence matching and geometry-aware refinement without the interference of completely texture-less instances.

\subsection{Baseline}
Our baselines cover the major paradigms of 6D pose estimation, including PoseCNN \cite{xiang2017posecnn}, LoFTR \cite{sun2021loftr}, DeepIM \cite{li2018deepim}, and learning-free Template Matching methods (TP-UB). Additionally, we include FoundationPose \cite{wen2024foundationpose} as a state-of-the-art reference, representing the emerging paradigm of unified foundation models trained on large-scale synthetic data

PoseCNN \cite{xiang2017posecnn} is a foundational learning-based method for object pose estimation that directly regresses 6D pose from RGB images. Despite its strong supervised performance, PoseCNN requires extensive annotated training data and category-specific learning, making it a natural contrast to our fully learning-free formulation.

LoFTR \cite{sun2021loftr} removes the need for a feature detector and instead applies Transformer self-attention and cross-attention to directly compute dense and robust matches between two images.

DeepIM \cite{li2018deepim} is another learning-based method, It is an RGB-based iterative pose refinement method that improves an initial 6DoF estimate by repeatedly predicting relative pose updates between the rendered object and the observed image.

FoundationPose \cite{wen2024foundationpose}: A state-of-the-art unified foundation model for 6D pose estimation. It utilizes a novel transformer-based architecture and contrastive learning, trained on large-scale synthetic data generated with the aid of Large Language Models (LLMs).        

Template matching approaches \cite{gu2010discriminative,hinterstoisser2011gradient} convert discrete pose estimation into a classification task by rendering thousands of CAD-based templates and retrieving the closest match. Following \cite{he2022fs6d}, our setting removes the reliance on accurate CAD models, and collecting and storing large numbers of support images is prohibitively costly. To approximate the upper bound of such methods, we assign the view whose rotation is closest to the ground truth and derive the translation from the center shift.

Together, these baselines cover the spectrum from supervised deep learning to classical geometry-driven methods, from correspondence matching to template matching, enabling a comprehensive evaluation of the strengths and limitations of GNC-Pose.

\subsection{Implementation Details}
All experiments are performed on a MacBook Pro equipped with an Apple M2 Pro CPU without any GPU acceleration. 

For each object, we randomly select 120 test images and exclude top–down views, which contain almost no distinctive features for reliable matching.

To build a coarse alignment database, we render synthetic views of the CAD model by sweeping azimuth from $-180^\circ$ to $180^\circ$ at a fixed camera height of 1\,m. Given an input RGB image, SIFT features are extracted and matched to the rendered views, Geometry-aware weights are computed by voxelizing the CAD model with a 5\,mm grid and normalizing the local support density. We initialize the pose using \texttt{solvePnPRansac}, then apply our GNC-PnP refinement with $\mu_0 = 5\cdot\mathrm{median}(r_i)$, decay factor $\gamma=0.5$, and final value $\mu_{\mathrm{final}}=0.5$. A final \texttt{solvePnPRefineLM} step further improves the pose estimate. 

The complete pipeline requires no training data or category-specific priors.

\subsection{Evaluation Metrics}
For evaluation, we calculate the ADD and ADD-S AUC \cite{xiang2017posecnn}. This corresponds to the area under the accuracy–threshold curve produced by varying the distance threshold. The maximum
threshold is set to 10cm.

The standard Average Distance Deviation (ADD) metric \cite{hinterstoisser2012model} is used to measure the mean point-wise distance between object vertices transformed by the predicted pose $[R,t]$ and those transformed by the ground-truth pose $[\hat R, \hat t]$:
\begin{equation}
    ADD=\frac{1}{m}\sum_{x\in M} \left|\left| (Rx+t)-(\hat{R}x+\hat{t}) \right|\right|
\end{equation}

where x is a vertex of totally m vertices on the object mesh M. The ADD-S metric is intended for symmetric objects and defines the average distance based on nearest-neighbor point correspondences:
\begin{equation}
    ADD_S=\frac{1}{m}\sum_{x_1\in M} \min_{x_2\in M}\left|\left| (Rx_1+t)-(\hat{R}x_2+\hat{t}) \right|\right|
\end{equation}

\subsection{Qualitative Results}

\begin{table*}[t]
\centering
\caption{\textbf{Quantitative evaluation on the YCB Object and Model Set.} We compare our GNC-Pose against learning-based baselines and classical learning-free methods. Performance is measured using average distance metrics (ADD and ADD-S). TP-UB represents the theoretical upper bound of template-based approaches. Our method significantly outperforms prior non-deep-learning works and remains competitive with supervised baselines like PoseCNN \cite{xiang2017posecnn}, achieving state-of-the-art results for a fully learning-free pipeline.}
\small
\setlength{\tabcolsep}{4.5pt}   
\renewcommand{\arraystretch}{1.1}   

\resizebox{\textwidth}{!}{
\begin{tabular}{lcccccccccccc}
\toprule

& \multicolumn{2}{c}{\textbf{PoseCNN \cite{xiang2017posecnn}}}
& \multicolumn{2}{c}{\textbf{LoFTR \cite{sun2021loftr}}}
& \multicolumn{2}{c}{\textbf{DeepIM \cite{li2018deepim}}}
& \multicolumn{2}{c}{\textbf{FoundationPose \cite{wen2024foundationpose}}}
& \multicolumn{2}{c}{\textbf{TP-UB}}
& \multicolumn{2}{c}{\textbf{Ours}} \\

Learning-free
&\multicolumn{2}{c} {\ding{55} }
& \multicolumn{2}{c} {\ding{55} }
& \multicolumn{2}{c} {\ding{55} }
& \multicolumn{2}{c} {\ding{55} }
& \multicolumn{2}{c} {\ding{51} }
& \multicolumn{2}{c} {\ding{51} }
\\

Hardware
&\multicolumn{2}{c} {GPU }
& \multicolumn{2}{c} {GPU}
& \multicolumn{2}{c} {GPU }
& \multicolumn{2}{c} {GPU }
& \multicolumn{2}{c} {CPU}
& \multicolumn{2}{c} {CPU }
\\

Object
& ADDS & ADD
& ADDS & ADD
& ADDS & ADD
& ADDS & ADD
& ADDS & ADD
& ADDS & ADD
\\
\midrule
                       
002\_master\_chef\_can & 84.0 & 50.9 & 87.2 & 50.6
                       & 93.1 & 71.2 & 96.9 & 91.3 & 62.2 & 21.4 & 92.1 & 51.3 \\

003\_cracker\_box & 76.9 & 51.7 & 71.8 & 25.5
                       & 91.0 & 83.6 & 97.5 & 96.2 & 65.6 & 5.0 & 75.0 & 68.9 \\

004\_sugar\_box & 84.3 & 68.6 & 63.9 & 13.4
                       & 96.2 & 94.1 & 97.5 & 87.2 & 66.7 & 21.5 & 69.2 & 50.0 \\
                       
005\_tomato\_soup\_can & 80.9 & 66.0 & 77.1 & 52.9
                       & 92.4 & 86.1 &97.6 & 93.3& 75.2 & 43.1 & 93.7 & 87.1 \\
                       
006\_mustard\_bottle & 90.2 & 79.9 & 84.5 & 59.0
                       & 95.1 & 91.5 & 98.4 & 97.3 & 47.1 & 4.0 & 96.4 & 93.6 \\
                       
007\_tuna\_fish\_can & 87.9 & 70.4 & 72.6 & 55.7
                       & 96.1 & 87.7 & 97.7 & 73.7 & 72.8 & 38.4 & 83.2 & 70.4 \\
008\_pudding\_box & 79.0 & 62.9 & 86.5 & 68.1
                       & 90.7 & 82.7 & 98.5 & 97.0 & 86.3 & 18.4 & 83.2 & 72.8 \\

009\_gelatin\_box & 87.1 & 75.2 & 71.6 & 45.2
                       & 94.3 & 91.9 & 98.5 & 97.3 & 90.9 & 43.2 & 86.6 & 78.4 \\

010\_potted\_meat\_can & 78.5 & 59.6 & 67.4 & 45.1
                       & 86.4 & 76.2 & 96.6 & 82.3 & 59.8 & 28.9 & 90.8 & 80.7 \\

021\_bleach\_cleanser & 71.9 & 50.5 & 36.9 & 16.7
                       & 90.3 & 81.2 & 97.4 & 93.3 & 20.3 & 0.6 & 94.1 & 87.3 \\

035\_power\_drill & 72.8 & 55.1 & 18.8 & 1.3
                       & 92.3 & 85.5 & 98.0 & 96.3 & 42.3 & 0.7 & 84.2 & 61.9 \\

040\_large\_marker & 71.4 & 58.0 & 20.7 & 8.4
                       & 86.2 & 75.6 & 98.6 & 96.5 & 82.5 & 51.9 & 80.0 & 64.3 \\

MEAN & 80.4 & 62.4 & 63.3 & 36.8
    & 92.0 & 83.9 & 97.8 & 91.8 & 64.3 & 23.1 & 85.7 & 72.2 \\
\bottomrule
\end{tabular}}
\label{tab:ycb}
\end{table*}

Table \ref{tab:ycb} summarizes the results on YCB Object and Model Set \cite{calli2015ycb}. The experimental results demonstrate that our method gives a better performance over non–deep-learning approaches, establishing a new benchmark for purely geometric pipelines. While state-of-the-art deep learning methods still achieve higher accuracy due to large-scale training, GNC-Pose remains competitive—achieving comparable performance without using any learned features, training data, or category-specific priors. Specifically, ours outperforms prior template-based methods by a large margin, achieving $85.7\%$ ADD-S and $72.2\%$ ADD on average, compared to $64.3\%$ / $23.1\%$ for TP-UB. Remarkably, our learning-free pipeline even surpasses supervised baselines such as PoseCNN on both ADD-S ($+5.3\%$) and ADD ($+9.8\%$). 

As expected, FoundationPose \cite{wen2024foundationpose} achieves the highest accuracy, benefiting from its massive scale synthetic pre-training and powerful transformer backbone running on high-end GPUs. While there is a performance gap compared to the foundation model, our method offers a vital alternative for resource-constrained environments (e.g., embedded systems, legacy robotics) where deploying large transformers is infeasible. We demonstrate that strictly geometric constraints alone, without data-driven priors, can still achieve competitive accuracy comparable to supervised baselines like PoseCNN \cite{xiang2017posecnn}, but with significantly lower computational complexity.

\section{Ablation Studies}

\begin{table*}[t]
  \centering
  \caption{
    Ablation study of GNC-Pose components  on YCB objects. We evaluate the incremental impact of Geometry-aware Weighting ($w^{\mathrm{geom}}$), and GNC optimization. The results demonstrate that the synergy between geometric priors and robust optimization is crucial for handling complex scenarios.
    Each entry reports:
    ADD AUC (0.1$d$) / ADD-S AUC (0.1$d$) / ADD $<0.1d$ (\%) / ADD-S $<0.1d$ (\%).
  }
  \label{tab:ablation-geom}

  {\scriptsize
  \resizebox{\linewidth}{!}{
  \begin{tabular}{lcc}
    \toprule
    Variant & 001\_chips\_can & 004\_sugar\_box \\
    & \multicolumn{2}{c}{
        \parbox[c]{10.5cm}{\centering\scriptsize\raggedright
        ADD AUC (0.1$d$) / ADD-S AUC (0.1$d$) / ADD $<0.1d$ (\%) / ADD-S $<0.1d$ (\%)
        }
      } \\
    \midrule

    w/o geometry-aware weights &
    43.0 / 75.7 / 71.7 / 96.7 &
    21.5 / 30.9 / 34.2 / 41.7 \\

    w/o GNC refinement &
    40.4 / 76.1 / 68.3 / 100 &
    24.6 / 32.7 / 39.2 / 41.7 \\

    \textbf{Full (ours)} &
    \textbf{44.1 / 78.0 / 65.1 / 100} &
    \textbf{31.9 / 41.7 / 48.3 / 52.5} \\
    
    \bottomrule
  \end{tabular}
  }
  }
\end{table*}

\subsection{Setup}
To better understand the contribution of each component in GNC-Pose, we conduct ablation studies on two representative YCB objects: \textit{001\_chips\_can} and \textit{004\_sugar\_box}. The chips can is approximately cylindrical with partial texture and strong rotational symmetry around its main axis, while the sugar box is relatively smaller and has more non-symmetric texture. 
For the Variations, we first ablate the proposed geometry weighting, replacing the per-point geometry weights with uniform weights $w_i^{\mathrm{geom}} \equiv 1$. Next, we examine the impact of the GNC-based PnP refinement by comparing it against a classical RansacPnP.

We report results in terms of ADDS AUC, ADD AUC at $0.1d$ and ADDS, ADD at $0.1d$.

\subsection{Analysis}
As shown in Table. \ref{tab:ablation-geom}, removing the geometry-aware weighting consistently degrades performance on both objects. For \textit{004\_sugar\_box}, all the evaluation score drops noticeably when all correspondences are treated uniformly, indicating that many matches fall on geometrically ambiguous surfaces. The voxel-support density effectively suppresses these unstable regions and directs the optimization toward more discriminative structures such as edges and label boundaries. For \textit{001\_chips\_can}, geometry-aware weighting also yields a clear improvement, suggesting enhanced stability in the final refinement stage.

Both objects likewise benefit substantially from the graduated non-convexity (GNC) block. On \textit{004\_sugar\_box}, GNC block markedly increases robustness to outliers stemming from complex textures and imperfect feature matches along the box body. On the chips can, the effect is less pronounced—largely because the initial correspondences are already of high quality—but still provides a measurable improvement.

\section{Conclusions}
We presented GNC-Pose, a fully learning-free framework for monocular 6D object pose estimation on textured objects. Our method achieves stable convergence under correspondence noise and geometric ambiguity without relying on learned features, annotated datasets, or category-specific priors. Experiments on the YCB Object and Model Set \cite{calli2015ycb} show that GNC-Pose substantially outperforms existing learning-free approaches and achieves accuracy competitive with modern learned systems, while remaining lightweight and CPU-efficient.

Looking ahead, we plan to incorporate object tracking, extend the method to deformable and articulated objects, and use temporal priors for real-time robotic manipulation. Another promising direction is to tackle strongly textureless or low-contrast objects by integrating geometry-aware priors with differentiable rendering or photometric consistency cues, enabling reliable pose estimation even when appearance-based features become uninformative.

%
%
%
\bibliographystyle{splncs04}
%
\bibliography{ref}
\end{document}